\documentclass[letterpaper, 10 pt, conference]{ieeeconf}
\IEEEoverridecommandlockouts                               
\usepackage{breqn}
\usepackage{cuted}
\usepackage{xcolor}
\usepackage{capt-of}
\usepackage{amsfonts}
\usepackage{hyperref}
\usepackage{multirow}
\usepackage{mathtools}
\usepackage{dblfloatfix}
\usepackage{makecell}
\usepackage{siunitx}
\usepackage{amsmath}
\usepackage{amsfonts}
\usepackage{amssymb}
\usepackage{graphicx}
\usepackage{algorithm}
\usepackage{algpseudocode}
\usepackage{booktabs}

\usepackage[font=small, labelfont=bf]{caption}

\usepackage{amssymb}

\usepackage{fancyhdr} 
\def\papername {CT-SGAL-LIO}

\renewcommand{\arraystretch}{1.2}
\definecolor{pink}{RGB}{255, 192, 203}

\newcommand{\del}[1]{}

\usepackage[T1]{fontenc}      % font encoding
\usepackage[utf8]{inputenc}   % input encoding
\usepackage[vietnamese]{babel}

\title{\LARGE \papername: Causal Continuous-Time Refinement of Discrete LiDAR–Inertial Odometry}

\author{%
    Nam Tran$^{1,\dagger}$,
    Thu Tran$^{1,\dagger}$,
    Toan Nguyen$^{2}$,
    William~J.~Beksi$^{3}$,
    and~Tuan~Dang$^{1,*}$%
    \thanks{$^{1}$Department of Electrical Engineering and Computer Science, University of Arkansas, Fayetteville, AR, USA; $^{2}$VinRobotics; $^{3}$Department of Computer Science, University of Texas at Dallas, Richardson, TX, USA. $^{\dagger}$Contributed to this work as interns at the Cognitive Robotics Lab, University of Arkansas. $^{*}$Corresponding author: Tuan Dang (\texttt{tuand@uark.edu}).}%
}

\begin{document}

\maketitle 

\thispagestyle{empty}
\pagestyle{empty}

\begin{abstract}

State-of-the-art LiDAR-inertial odometry methods, including those on richer state manifolds, achieve strong global accuracy, yet their pose stream alternates between propagation and correction and can oscillate over short horizons even when globally accurate. Most existing methods follow these approaches: they either modify the filter to enable faster updates or estimate a continuous-time trajectory from raw LiDAR and inertial data. The latter involves extra optimization, delaying state updates and becoming less robust during rapid rotations or accelerations. We propose \papername, a continuous-time refinement module that smooths the posterior pose stream of a discrete $\mathrm{SGal}(3)$ estimator without modifying the estimator itself. The key novelty is to perform continuous-time trajectory refinement directly on the projected $\mathrm{SGal}(3)$ posterior: we fit a cubic trajectory over a bounded window of recent poses and evaluate it at a small fixed lag. This places the query time within the observed interval, so the refined pose is obtained by interpolation rather than extrapolation, while preserving causal operation. CT-SGAL-LIO improves relative trajectory accuracy on 17 sequences from three public benchmarks. Across 17 sequences from GRAND-TOUR, MCD, and GEODE, \papername{} improves short-term motion accuracy while largely preserving global accuracy. RTE improves by up to 70\% on GRAND-TOUR and on all seven GEODE sequences, with 9.9--51.1\% reductions. A 19.5-min R-Campus run also reduces posterior jumps by 85.4\% with only a 4.7 mm change in the end-to-start gap. We plan to release our source code and results to support community validation and reproducibility.

% The largest gain reaches 70\% in relative error, with little change in global accuracy on most sequences. On the hardest sequence, absolute error drops by more than 95\% and relative error by nearly 80\%. We plan to release our source code and results to support community validation and reproducibility.

% \cm{We need \textbf{VIDEO} too}

\end{abstract}

\section{Introduction}

% \cm{@Toan: Put your text here as soon as you can since we are almost out of time.}

\begin{figure}[t]
    \centering
    \includegraphics[width=0.49\textwidth]{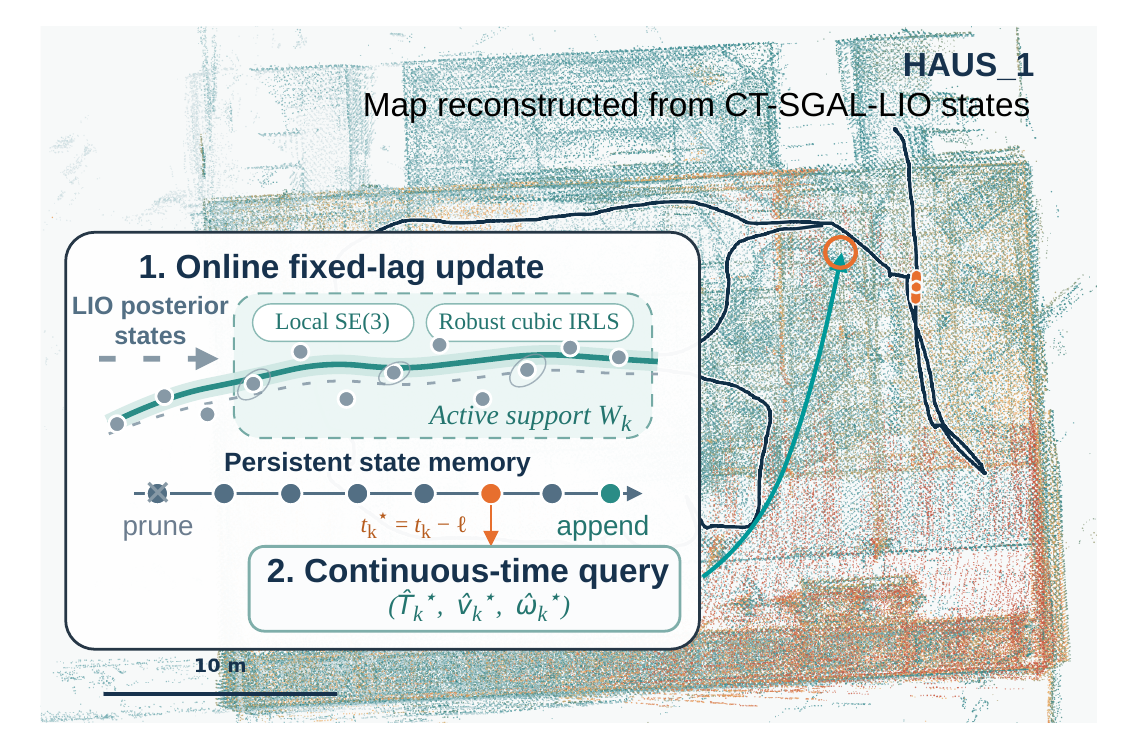}
    \caption{\papername{} maintains a persistent fixed-lag memory of incoming LIO posterior states and reconstructs a continuous trajectory by robust metric-weighted cubic fitting in a local $SE(3)$ tangent chart. At the delayed query time $t_k^{\star}=t_k-\ell$, the fitted trajectory provides the refined pose and its linear and angular velocities. The orange segment marks the illustrated local support. }
    \label{fig:ctsgal_concept}
\end{figure}

LiDAR--inertial odometry (LIO) underlies mapping, planning, and control on mobile robots, and downstream consumers routinely query the pose stream at rates decoupled from the LiDAR scan rate.
State-of-the-art tightly coupled iterated filters, most recently formulated on the Galilean manifold $\mathrm{SGal}(3)$~\cite{perez2025limoncello}, achieve strong global trajectory precision.
However, the state stream that such filters publish alternates between IMU-propagated states and LiDAR-corrected states, which can exhibit different local error characteristics.
As a result, a globally accurate trajectory can still contain short-horizon oscillations, visible even on our own baseline: on the GRAND-TOUR~\cite{frey2026grandtour} \textit{HAUS\_1} sequence, the discrete estimator we build on already achieves a low global error, yet its relative motion error over short intervals still fluctuates, and \papername~reduces this short-term error by 67\% without altering the global trajectory (Fig.~\ref{fig:ctsgal_concept}).

Existing remedies for this local inconsistency modify the discrete estimator itself. One direction raises the state-update rate by restructuring the filter to process individual points rather than whole scans. A second direction replaces the discrete state altogether with a continuous-time trajectory representation fitted jointly to raw LiDAR and, where available, inertial residuals. This approach works well, but it has two drawbacks, discussed in Section~\ref{sec:related_work}. Smooth parametric trajectories can struggle with aggressive motion, and fitting them with future measurements introduces delay. Neither direction considers a simpler alternative: keep the discrete LIO estimator and its map untouched, and refine only the pose stream after it is produced.

We answer this question with \papername, an online continuous-time refinement layer that treats the discrete posterior as an observation rather than as a state to be re-derived.
\papername~maintains a persistent, bounded-memory buffer of recent posterior poses and fits a robust cubic trajectory over this window in a local $\mathrm{SE}(3)$ Lie-algebra chart.
The refined trajectory is evaluated at a fixed lag, placing the query time inside the observed window rather than at its boundary. This lets us obtain the pose by interpolation rather than extrapolation, without sacrificing causality. Because \papername~consumes only the timestamped $\mathrm{SE}(3)$ projection of the estimator's posterior state and feeds nothing back into the filter or its map, it is decoupled from the underlying estimator and can in principle attach to any discrete LiDAR-inertial system that exposes a pose stream. In this paper, we instantiate it on top of LIMOncello~\cite{perez2025limoncello}, the estimator whose $\mathrm{SGal}(3)$ posterior it refines.

Our contributions are as follows.
\begin{itemize}
    \item We propose a model-agnostic, causal, fixed-lag continuous-time refinement layer that operates purely on the posterior pose stream of a discrete LIO estimator, without coupling to raw sensor residuals or to the map.
    \item We formulate this refinement as robust cubic trajectory fitting in a local $\mathrm{SE}(3)$ Lie-algebra chart, combining iteratively reweighted least squares with a curvature prior tailored to short, bounded temporal windows.
    \item We show that attaching \papername~to LIMOncello's
$\mathrm{SGal}(3)$ posterior consistently improves short-term motion
accuracy while generally preserving global trajectory accuracy across
17 sequences from three public benchmarks (GRAND-TOUR, MCD~\cite{nguyen2024mcd},
and GEODE~\cite{chen2026heterogeneous}). Specifically, \papername~
reduces RTE by up to 70\% on GRAND-TOUR and by 9.9--51.1\% on all
seven GEODE sequences, while ranking first or second in RTE on all
six evaluated MCD sequences.
    % \item We analyze the conditions under which the refinement helps versus provides no benefit, showing that its gain tracks the amount of short-term fluctuation already present in the underlying posterior.
\end{itemize}

The remainder of this paper is organized as follows.
Section~\ref{sec:related_work} reviews discrete-time LIO, manifold state representations, continuous-time trajectory estimation, and fixed-lag smoothing.
Section~\ref{sec:system} overviews the system architecture.
Section~\ref{sec:methodology} details the fixed-lag continuous-time refinement formulation.
Section~\ref{sec:evaluation} presents our experimental evaluation.
Section~\ref{sec:conclusions} concludes the paper.
\section{Related Work}
\label{sec:related_work}

% \cm{@Toan: Put your text here as soon as you can since we are almost out of time.}

\subsection{Discrete-time LIO on Manifold State Representations}

Most LiDAR odometry systems follow LOAM~\cite{zhang2014loam}: extract geometric features, estimate scan-to-pose correspondences, and register to a map. Tightly coupled LiDAR--inertial designs then became standard, either as sliding-window smoothers~\cite{shan2020lio} or as iterated filters standardized by IKFoM~\cite{he2023symbolic}. FAST-LIO~\cite{xu2021fast} introduced IMU back-propagation with a Kalman gain that scales with the state dimension rather than the measurement dimension, and FAST-LIO2~\cite{xu2022fast} removed hand-engineered feature extraction by registering raw points directly to an incremental $k$-d tree, ikd-Tree~\cite{cai2021ikd}. This family publishes only one pose per scan, and Point-LIO~\cite{he2023point, tran2026hp2slam} shows that this scan-rate update imposes a Nyquist-limited bandwidth. The state representation from this work is typically a bundle manifold $SO(3)\times\mathbb{R}^{6}$, which ignores the coupling between rotation and translation; richer Lie groups restore this coupling along the nested hierarchy $SO(3)<SE(3)<SE_2(3)<SGal(3)$~\cite{kelly2023all}, with $SE_2(3)$ underpinning invariant filtering~\cite{barrau2016invariant} and $SGal(3)$ additionally carrying a temporal coordinate. LIMOncello~\cite{perez2025limoncello} is, to our knowledge, the first tightly coupled LIO system on $SGal(3)$ within IKFoM, and is the estimator we build on: because $SE(3)$ is a proper subgroup of $SGal(3)$, the pose $\Pi_{SE(3)}(G_i)$ is a well-defined projection, and it is the quantity on which our refinement layer operates. We leave this discrete-time estimator unchanged and instead regularize the state sequence that it already produces.

\subsection{Continuous-time trajectory estimation}

Continuous-time methods represent the trajectory as a function of time, so measurements can be fused at their true timestamps and motion distortion is estimated jointly with the motion~\cite{talbot2025continuous}. Two parameterizations are commonly used. Piecewise-linear models assume constant velocity over short intervals, as in standard deskewing. Cubic B-splines use more parameters but give a smoother trajectory. CT-ICP~\cite{dellenbach2022ct} uses a begin and an end pose per scan and deliberately allows inter-scan discontinuity, while Traj-LO~\cite{zheng2024traj} regularises short linear segments. Cubic B-splines are used in sliding-window LiDAR--inertial fusion by CLIC and SLICT2~\cite{lv2023continuous,nguyen2024eigen} and, more recently, inside a recursive filter, RESPLE~\cite{cao2025resple}, which embeds spline control points directly in an iterated Kalman filter and avoids sliding-window optimization altogether. DLIO~\cite{chen2022direct} instead sidesteps control-point fitting entirely, integrating a closed-form constant-jerk model. Smooth trajectory methods like CT-ICP, DLIO, and Point-LIO argue that control-point fitting can smooth out rapid motion, while RESPLE finds cubic splines sufficiently flexible in practice. Continuous-time methods also introduce latency because future measurements must be collected before the trajectory parameters are well constrained and the state can be queried~\cite{he2023point}. CT-ICP, CLIC, SLICT2, and RESPLE  optimize their trajectory parameters directly against raw LiDAR (and, where used, inertial) residuals, so the continuous-time representation \emph{is} their estimator rather than a separable layer on top of it. \papername~instead operates only on the timestamped pose stream that LIMOncello already publishes, with no feedback into the filter or its map, which lets the refinement layer attach to any discrete LiDAR-inertial estimator that exposes a pose stream.

\subsection{Smoothing and Fixed-Lag Estimation in LiDAR Odometry}

Fixed-lag smoothing bounds the estimation delay by withholding the output until a short buffer of subsequent measurements becomes available~\cite{dong2011motion}. In LiDAR odometry, smoothing is usually applied instead as a factor-graph back-end~\cite{shan2020lio} or a sliding window of spline control points with marginalisation ~\cite{lv2023continuous,nguyen2024eigen,zheng2024traj}. A related formulation is Wildcat~\cite{ramezani2022wildcat}, which fits a cubic B-spline to sample poses by IRLS, following the offline systems of Bosse and Zlot~\cite{bosse2012zebedee}; there, however, the spline is internal to the estimator, since the sample poses are themselves optimized against IMU and surfel residuals and the interpolated trajectory is fed back to update correspondences. Because Wildcat fits the spline inside its pipeline, the spline is part of the estimator rather than a separate refinement stage. Comparing our method directly with Wildcat would therefore confound the effect of trajectory refinement with differences in the underlying estimator. To isolate the effect of our refinement, we compare against LIMOncello and RESPLE, the discrete estimators whose posterior pose streams serve as input to our method. \papername{} treats these posterior poses as observations, fits a continuous trajectory over a bounded history window, and evaluates the trajectory at a fixed lag so that the query time lies inside the observed interval. This keeps the refinement causal while leaving the original LiDAR--inertial estimator and map unchanged.
\begin{figure*}[th]
    \centering
    \includegraphics[width=1.0\textwidth]{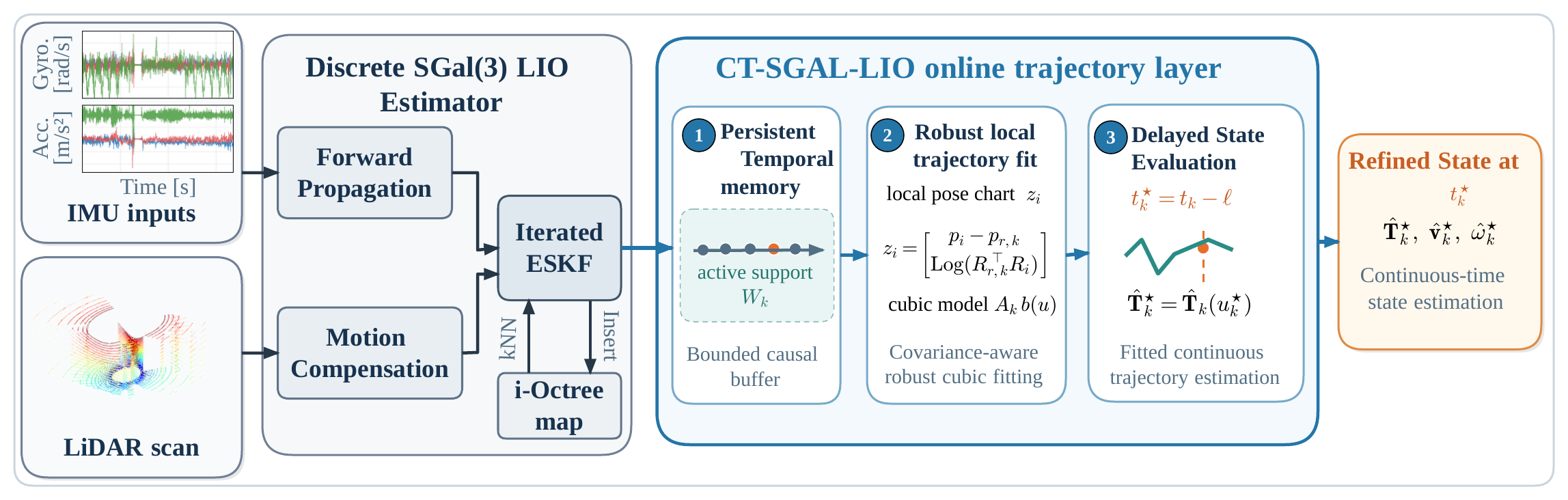}
    \caption{Overview of \papername{}. Posterior states from the discrete SGal(3) LIO estimator are maintained in a bounded temporal memory and reconstructed by covariance-aware robust cubic fitting. The resulting
continuous trajectory is evaluated at the delayed query time
$t_k^\star=t_k-\ell$ to obtain the refined state.}
    \label{fig:ct_sgal_system}
\end{figure*}

\section{System Overview}
\label{sec:system}

% \cm{@Nam + Thu: Describe this section. Basically, it is derived directly from the pipeline} \\

% The overall architecture of CT-SGAL-LIO is illustrated in
% Fig.~\ref{fig:ct_sgal_system}. A discrete SGal(3)-based LiDAR--inertial
% estimator provides high-rate posterior states after IMU propagation and
% LiDAR correction. Rather than treating these posterior states as independent
% discrete estimates, CT-SGAL maintains a persistent fixed-lag representation
% of their recent temporal evolution and reconstructs a locally continuous
% trajectory directly in posterior space.

% Within the active window $\mathcal W_k$, posterior poses are first expressed
% in a local tangent chart together with their associated uncertainty.
% CT-SGAL then estimates a robust cubic trajectory using covariance-conditioned
% observations and temporal regularization. The reconstructed trajectory is
% evaluated at the interior delayed time
% $t_k^\star=t_k-\ell$ and mapped back to the pose manifold to obtain the
% refined pose and motion state.

% This formulation decouples continuous-time refinement from the LiDAR
% measurement and mapping front-end while retaining a causal online structure.
% Since only a bounded temporal support is optimized and the trajectory model
% has fixed dimension, CT-SGAL provides persistent continuous-time refinement
% without introducing a growing trajectory optimization problem.

The overall architecture of CT-SGAL-LIO is illustrated in Fig.~\ref{fig:ct_sgal_system}. A discrete SGal(3)-based LiDAR--inertial estimator provides high-rate posterior states after IMU propagation and LiDAR correction. Instead of treating each posterior state independently, \papername{} keeps a short history of recent states and fits a continuous trajectory through them; \papername{} uses the recent posterior poses to fit a cubic trajectory that maps each pose to a local tangent space, and its covariance determines how strongly it influences the fit. A temporal regularizer limits unnecessary curvature. The trajectory is evaluated at the interior delayed time $t_k^\star=t_k-\ell$ and mapped back to the pose manifold to obtain the refined pose and motion state. This design keeps the continuous-time refinement module separate from the LiDAR front end while remaining causal and suitable for online use. 

\section{Methodology}
\label{sec:methodology}
\subsection{Problem Formulation}

\label{sec:problem_formulation}

We consider a LiDAR--inertial estimator whose posterior state at time
$t_i$ evolves on the special Galilean group,
\begin{equation}
\mathbf G_i =
\begin{bmatrix}
\mathbf R_i & \mathbf v_i & \mathbf p_i\\
\mathbf 0^\top & 1 & t_i\\
\mathbf 0^\top & 0 & 1
\end{bmatrix}
\in \mathrm{SGal}(3),
\label{eq:sgal_state}
\end{equation}
where $\mathbf R_i\in\mathrm{SO}(3)$, $\mathbf p_i\in\mathbb R^3$,
and $\mathbf v_i\in\mathbb R^3$ are the estimated orientation, position,
and velocity. We use LIMOncello~\cite{perez2025limoncello} as the discrete underlying
 estimator. CT-SGAL constructs a continuous-time readout from its
pose posterior
$\mathbf T_i=(\mathbf R_i,\mathbf p_i)\in\mathrm{SE}(3)$
and the associated covariance $\boldsymbol\Sigma_i$.

The estimator produces timestamped posterior states
$\{(t_i,\mathbf T_i,\boldsymbol\Sigma_i)\}_{i=1}^{k}$.
Because IMU propagation and LiDAR correction occur at different rates,
the sampled trajectory can contain short-term variations around correction
events. These variations become visible when the trajectory is queried at
a higher rate than the LiDAR updates.

\papername{} processes this stream online using a time-bounded pose buffer.
At the arrival of state $k$, the new posterior
$(t_k,\mathbf T_k,\boldsymbol\Sigma_k)$ is inserted into the buffer
$\mathcal M_k$, while states older than the memory horizon
$\Delta_m$ are removed by $\mathcal P_{\Delta_m}$. The output is placed
at the fixed-lag time $t_k^\star=t_k-\ell$, and the active window is
defined as
\begin{equation}
\mathcal W_k
=
\left\{
(t_i,\mathbf T_i,\boldsymbol\Sigma_i)\in\mathcal M_k
\;\middle|\;
t_k^\star-W\leq t_i\leq t_k
\right\}.
\label{eq:online_support}
\end{equation}
Here, $W$ is the history retained before the query time, and $\ell$
is the output lag. Unlike pairwise interpolation, each CT-SGAL estimate
uses all posterior states in $\mathcal W_k$, while $\mathcal M_k$ persists
across successive updates.

The active interval $[t_k^\star-W,t_k]$ is normalized before fitting.
Its center is
$c_k=(t_k^\star-W+t_k)/2$, with half-width
$h=(W+\ell)/2$. Each sample is represented by the normalized coordinate
\begin{equation}
u_i=\frac{t_i-c_k}{h},
\qquad
u_k^\star=\frac{t_k^\star-c_k}{h}
          =\frac{W-\ell}{W+\ell}.
\label{eq:normalized_time}
\end{equation}
This maps $[t_k^\star-W,t_k]$ to $[-1,1]$. For $W>0$ and $\ell>0$,
the query lies strictly inside the observed interval. The fit therefore
uses states both before and after $t_k^\star$ without accessing
measurements later than the current processing time $t_k$.

Directly fitting rotation matrices does not preserve $\mathrm{SO}(3)$.
The first pose in $\mathcal W_k$ serves as the local reference,
$\mathbf T_{r,k}=(\mathbf R_{r,k},\mathbf p_{r,k})$.
Each pose is represented by its translation and rotation relative to this
reference:
\begin{equation}
\mathbf z_i=
\begin{bmatrix}
\mathbf p_i-\mathbf p_{r,k}\\
\operatorname{Log}\!\left(
\mathbf R_{r,k}^{\top}\mathbf R_i
\right)
\end{bmatrix}
\in\mathbb R^6 .
\label{eq:local_pose_chart}
\end{equation}

The resulting six-dimensional coordinates are modeled by a cubic
polynomial. Using
$\mathbf b(u)=[1,u,u^2,u^3]^\top$ and
$\mathbf A_k=[\mathbf A_{p,k}^\top\ \mathbf A_{R,k}^\top]^\top$,
with
$\mathbf A_{p,k},\mathbf A_{R,k}\in\mathbb R^{3\times4}$,
the fitted trajectory is
\begin{equation}
\hat{\mathbf z}_k(u)
=
\mathbf A_k\mathbf b(u).
\label{eq:cubic_trajectory}
\end{equation}
The corresponding
continuous pose is
\begin{equation}
\hat{\mathbf T}_k(u)=
\begin{bmatrix}
\mathbf R_{r,k}
\operatorname{Exp}\!\left(\mathbf A_{R,k}\mathbf b(u)\right)
&
\mathbf p_{r,k}+\mathbf A_{p,k}\mathbf b(u)\\
\mathbf 0^\top & 1
\end{bmatrix}.
\label{eq:continuous_pose}
\end{equation}
The exponential map ensures that the reconstructed orientation remains in
$\mathrm{SO}(3)$ for every $u$.

Let $\mathcal I_k$ contain the indices of the observations in
$\mathcal W_k$, and define
$\mathbf r_i(\mathbf A)=\mathbf z_i-\mathbf A\mathbf b(u_i)$.
We estimate the polynomial coefficients by
\begin{equation}
\begin{aligned}
\mathbf A_k^\star
=\arg\min_{\mathbf A}\;&
\sum_{i\in\mathcal I_k}
\rho_\delta
\!\left(
\|\mathbf r_i(\mathbf A)\|_{\boldsymbol\Omega_i}
\right)\\
&+
\sum_{q\in\mathcal Q}
\left\|
\mathbf A\ddot{\mathbf b}(q)
\right\|_{\boldsymbol\Lambda}^{2}
+
\lambda_d\|\mathbf A\|_F^2 ,
\end{aligned}
\label{eq:ct_sgal_objective}
\end{equation}
where
$\|\mathbf x\|_{\boldsymbol\Omega}
=
\sqrt{\mathbf x^\top\boldsymbol\Omega\mathbf x}$,
$\rho_\delta$ is the Huber loss~\cite{huber1992robust}, and
$\mathcal Q=\{-1,-0.5,0,0.5,1\}$ contains the curvature evaluation
points. We use
$\boldsymbol\Lambda=
\operatorname{diag}(\lambda_p\mathbf I_3,\lambda_R\mathbf I_3)$
to tune translational and rotational curvature separately.

The first term fits the recent posterior states using their uncertainty.
The Huber loss limits the effect of abrupt correction events, while the
curvature term suppresses oscillatory solutions. Thus, each update combines
multiple propagation and correction states rather than refining the newest
pose independently.

\subsection{Online Estimation and Implementation Details}
\label{sec:online_implementation}

CT-SGAL maintains $\mathcal M_k$ as a time-ordered buffer. A state with the
same timestamp as the newest entry replaces that entry, while out-of-order
states are discarded. Fitting starts once the buffer spans the required
interval and is repeated whenever the query time $t_k^\star$ advances.

The input covariance controls the weight of each posterior pose. We define
a minimum covariance
$\boldsymbol{\Sigma}_0=
\operatorname{diag}(\sigma_p^2\mathbf I_3,\sigma_R^2\mathbf I_3)$.
Before inversion, a valid covariance is conditioned as
\begin{equation}
\begin{aligned}
\tilde{\boldsymbol{\Sigma}}_i
&=
\operatorname{EigFloor}\!\left(
\operatorname{sym}(\boldsymbol{\Sigma}_i)
+\boldsymbol{\Sigma}_0,\epsilon
\right),\\
\boldsymbol{\Omega}_i
&=
\tilde{\boldsymbol{\Sigma}}_i^{-1}.
\end{aligned}
\label{eq:covariance_weight}
\end{equation}
where
$\operatorname{sym}(\boldsymbol{\Sigma})
=(\boldsymbol{\Sigma}+\boldsymbol{\Sigma}^{\top})/2$,
and $\operatorname{EigFloor}(\cdot,\epsilon)$ replaces eigenvalues smaller
than $\epsilon$ by $\epsilon$. If the input covariance is unavailable or
invalid, we use
$\tilde{\boldsymbol{\Sigma}}_i=\boldsymbol{\Sigma}_0$.
Thus, posterior poses with lower uncertainty receive greater weight in the
fit.

The robust objective in Eq.~\eqref{eq:ct_sgal_objective} is solved using
iteratively reweighted least squares (IRLS). At iteration $j$, the residual
of observation $i$ is
$\mathbf r_i^{(j)}=\mathbf r_i(\mathbf A_k^{(j)})$.
Its covariance-normalized magnitude is
\begin{equation}
e_i^{(j)}
=
\sqrt{
\mathbf r_i^{(j)\top}
\boldsymbol\Omega_i
\mathbf r_i^{(j)}
}.
\label{eq:mahalanobis_residual}
\end{equation}
The corresponding Huber weight is
\begin{equation}
w_i^{(j)}
=
\begin{cases}
1, & e_i^{(j)}\leq\delta,\\[1mm]
\delta/e_i^{(j)}, & e_i^{(j)}>\delta.
\end{cases}
\label{eq:huber_weight}
\end{equation}
Thus, $\boldsymbol\Omega_i$ accounts for the uncertainty of each posterior
state, while $w_i^{(j)}$ reduces the influence of residuals that exceed the
Huber threshold.

With the robust weights fixed, the cubic coefficients form a
24-dimensional weighted least-squares problem. We vectorize
$\mathbf A_k$ as
$\mathbf a_k=\operatorname{vec}(\mathbf A_k^\top)\in\mathbb R^{24}$
and solve the normal system
\begin{equation}
\mathbf H_k^{(j)}
\mathbf a_k^{(j+1)}
=
\mathbf g_k^{(j)}.
\label{eq:irls_system}
\end{equation}
Here, $\mathbf H_k^{(j)}$ and $\mathbf g_k^{(j)}$ collect the weighted
data term, the curvature term, and the damping term in
Eq.~\eqref{eq:ct_sgal_objective}. Since the cubic trajectory has four
coefficients for each of the six pose coordinates, the system size remains
$24\times24$ regardless of the total trajectory length.

After the final IRLS iteration, the fitted trajectory is evaluated at the
fixed-lag coordinate $u_k^\star$. The refined pose is
\begin{equation}
\hat{\mathbf T}_k^\star
=
\hat{\mathbf T}_k(u_k^\star).
\label{eq:refined_pose}
\end{equation}
Because $u=(t-c_k)/h$, we have $du/dt=1/h$. The translational velocity
therefore follows directly from the derivative of the cubic basis:
\begin{equation}
\hat{\mathbf v}_k^\star
=
\mathbf A_{p,k}^\star
\dot{\mathbf b}(u_k^\star),
\qquad
\dot{\mathbf b}(u)
=
\frac{1}{h}
[0,1,2u,3u^2]^\top .
\label{eq:refined_velocity}
\end{equation}

For rotation, we use a centered Lie-group difference around the query time.
With $\Delta u=\Delta t/h$, the angular velocity is
\begin{equation}
\hat{\boldsymbol\omega}_k^\star
=
\frac{1}{2\Delta t}
\hat{\mathbf R}_k^\star
\operatorname{Log}\!\left(
\hat{\mathbf R}_k(u_k^\star-\Delta u)^\top
\hat{\mathbf R}_k(u_k^\star+\Delta u)
\right).
\label{eq:angular_velocity}
\end{equation}
The online output is
$(\hat{\mathbf T}_k^\star,
\hat{\mathbf v}_k^\star,
\hat{\boldsymbol\omega}_k^\star)$,
with all quantities obtained from the same fitted trajectory.
% \subsubsection{SO(3) Manifold}
% \subsubsection{SGal(3)}

\section{Evaluations}
\label{sec:evaluation}

In this section, we present an extensive experimental evaluation of
\papername, focusing on both global trajectory accuracy and short-term
motion consistency. The evaluation is conducted on GRAND-TOUR, MCD, and
GEODE, which collectively provide diverse sensing configurations, motion
patterns, and challenging geometric conditions. Our experiments investigate
whether the proposed fixed-lag continuous-time refinement can reduce local
trajectory oscillations while preserving the global accuracy of the underlying
discrete-time estimator. In addition to the benchmark evaluation, we use a long-duration
R-Campus deployment to assess the long-horizon stability of CT-SGAL
and its effect on the global structure of the underlying LIO trajectory.

\subsection{Experimental Setup}
We evaluated \papername~on a total of 17 sequences from three public datasets: four GRAND-TOUR sequences, six sequences from the NTU Campus subset of MCD, and seven challenging sequences from GEODE. The selected datasets also span different LiDAR--inertial sensing configurations: the GRAND-TOUR sequences use a Hesai XT32 LiDAR with a NovAtel CPT7 inertial system; for MCD NTU Campus we use the Livox Mid-70 LiDAR with a VectorNav VN100 IMU; and for GEODE we use the $\alpha$ platform, a Velodyne VLP-16 LiDAR with an Xsens MTi-30 IMU. Together, these datasets cover a wide range of sensing configurations, motion patterns, and environmental
conditions, including indoor and outdoor scenes, day and night operation,
unstructured terrain, and geometrically challenging scenarios.

\begin{figure}[h]
    \centering
    \includegraphics[width=0.49\textwidth]{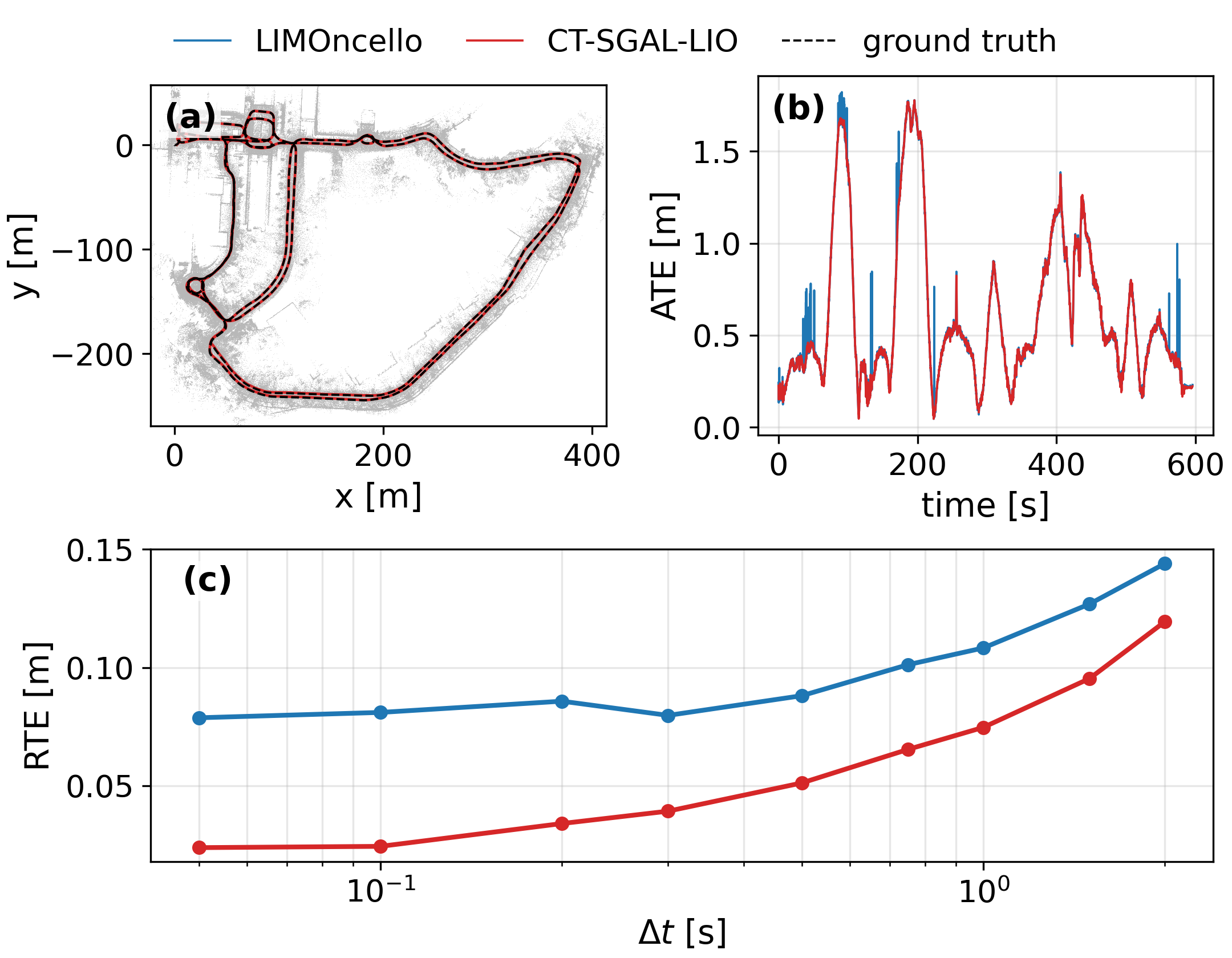}
    \caption{Qualitative comparison between LIMOncello and \papername~on the
    \textit{day01} sequence of the MCD NTU Campus dataset.
    (a) Estimated trajectories compared with the ground truth,
    (b) ATE over time, and
    (c) RTE evaluated at different time intervals $\Delta t$.
    \papername~preserves a trajectory close to the baseline while substantially
    reducing short-term relative motion error.}
    \label{fig:ctsgal_qualitative_day01}
\end{figure}

% \begin{figure}[h!]
%     \centering
%     \includegraphics[width=0.5\textwidth]{figures/enwide_katzensee_d_ct_sgal_qualitative.pdf}
%     \caption{Qualitative trajectory comparison on the ENWIDE Katzensee_d sequence. The main panel shows the aligned XY trajectories, while the inset magnifies a representative local segment. CT-SGAL-LIO closely follows the ground truth, with FAST-LIO2 also remaining accurate, whereas LIMOncello and RESPLE exhibit larger trajectory deviations.}
%     \label{fig:enwide_katzersee_qualitative}
% \end{figure}

\begin{figure}[h!]
    \centering
    \includegraphics[width=0.49\textwidth]    {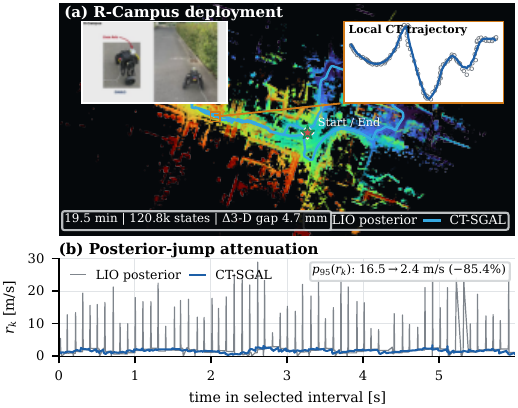}
    \caption{Long-duration online behavior of CT-SGAL on the R-Campus
    deployment. (a) The refined trajectory remains closely aligned with the
    LIO posterior over a 19.5-min sequence, while the inset shows a locally
    continuous CT trajectory reconstructed from discrete posterior samples.
    (b) CT-SGAL strongly attenuates consecutive-state posterior jumps, reducing
    $p_{95}(r_k)$ from $16.5$ to $2.4$\,m/s ($85.4\%$), while introducing only
    a $4.7$\,mm change in the 3-D end-to-start gap.}
    \label{fig:rcampus_qualitative}
\end{figure}

We compare \papername~with its discrete-time baseline LIMOncello, together
with RESPLE and FAST-LIO2. Trajectory accuracy is evaluated using the absolute
trajectory error (ATE) RMSE and the relative trajectory error (RTE) RMSE at a
temporal interval of $0.1\,\mathrm{s}$, following standard trajectory evaluation
practice~\cite{zhang2018tutorial} and computed with the \texttt{evo}
toolbox~\cite{grupp2017evo}. Estimated trajectories are synchronized
with the corresponding ground truth, and ATE is computed after a global rigid
$\mathrm{SE}(3)$ alignment. ATE measures global trajectory accuracy, whereas RTE
characterizes short-term relative motion consistency and is therefore particularly
relevant for evaluating the proposed continuous-time refinement. Lower values
indicate better performance for both metrics.

\subsection{Qualitative Results}

Figure~\ref{fig:ctsgal_qualitative_day01} illustrates the effect of \papername{}
on the MCD \textit{day01} sequence. While the refined and baseline trajectories
remain nearly identical at the global scale, \papername{} consistently reduces
the relative motion error over short evaluation intervals. This indicates that
the proposed fixed-lag reconstruction primarily removes high-frequency
trajectory fluctuations without noticeably altering the underlying global
estimate.

\begin{table*}[h]
    \centering
    \caption{Quantitative comparison on the GRAND-TOUR dataset in terms of 
    ATE RMSE and RTE RMSE at a time interval of $0.1$\,s.
    All values are reported in meters. Lower is better, and the best result
    for each metric is highlighted in \textbf{bold}.}
    \label{tab:grandtour_results}

    \small
    \setlength{\tabcolsep}{6pt}
    \renewcommand{\arraystretch}{1.15}

    \begin{tabular}{l cc cc cc cc}
        \toprule
        \multirow{2}{*}{Method}
        & \multicolumn{2}{c}{ARC1}
        & \multicolumn{2}{c}{Con\_1}
        & \multicolumn{2}{c}{Con\_3}
        & \multicolumn{2}{c}{HAUS\_1} \\

        \cmidrule(lr){2-3}
        \cmidrule(lr){4-5}
        \cmidrule(lr){6-7}
        \cmidrule(lr){8-9}

        & ATE & RTE$_{0.1s}$
        & ATE & RTE$_{0.1s}$
        & ATE & RTE$_{0.1s}$
        & ATE & RTE$_{0.1s}$ \\
        \midrule

        LIMOncello
        & 0.0686 & 0.0296
        & 0.0678 & 0.0310
        & 0.0236 & 0.0158
        & 0.0277 & 0.0257 \\

        RESPLE
        & 0.0251 & 0.0305
        & 0.0630 & 0.0343
        & 0.0635 & 0.0236
        & 0.0393 & 0.0307 \\

        FAST-LIO2
        & \textbf{0.0250} & 0.0284
        & \textbf{0.0311} & 0.0321
        & 15.6995 & 0.0843
        & 0.0214 & 0.0285 \\

        \midrule

        \papername
        & 0.0654 & \textbf{0.0123}
        & 0.0635 & \textbf{0.0090}
        & \textbf{0.0208} & \textbf{0.0085}
        & \textbf{0.0189} & \textbf{0.0084} \\

        \bottomrule
    \end{tabular}
\end{table*}

\begin{table*}[h]
    \centering
    \caption{Quantitative comparison on the NTU Campus subset of the MCD dataset.
    We evaluate day and night sequences recorded using a Livox Mid70 LiDAR and a VN100 IMU.
    ATE RMSE and RTE RMSE at a time interval of $0.1$\,s are reported in meters.
    Lower is better. The best and second-best results for each metric are highlighted
    in \textbf{bold} and \underline{underlined}, respectively.}
    \label{tab:mcd_results}

    \setlength{\tabcolsep}{5pt}
    \renewcommand{\arraystretch}{1.15}

    \resizebox{\textwidth}{!}{%
    \begin{tabular}{l cc cc cc cc cc cc}
        \toprule
        \multirow{2}{*}{Method}
        & \multicolumn{2}{c}{day01}
        & \multicolumn{2}{c}{day02}
        & \multicolumn{2}{c}{day10}
        & \multicolumn{2}{c}{night04}
        & \multicolumn{2}{c}{night08}
        & \multicolumn{2}{c}{night13} \\

        \cmidrule(lr){2-3}
        \cmidrule(lr){4-5}
        \cmidrule(lr){6-7}
        \cmidrule(lr){8-9}
        \cmidrule(lr){10-11}
        \cmidrule(lr){12-13}

        & ATE & RTE$_{0.1s}$
        & ATE & RTE$_{0.1s}$
        & ATE & RTE$_{0.1s}$
        & ATE & RTE$_{0.1s}$
        & ATE & RTE$_{0.1s}$
        & ATE & RTE$_{0.1s}$ \\
        \midrule

        LIMOncello
        & 0.7416 & 0.0810
        & \underline{0.2246} & \underline{0.0185}
        & 3.1404 & \textbf{0.0312}
        & \textbf{0.8877} & \underline{0.0221}
        & \underline{1.0046} & \textbf{0.0232}
        & 1.6388 & 0.0690 \\

        RESPLE
        & \textbf{0.4693} & 0.0276
        & 0.2634 & 0.1712
        & \underline{1.6488} & 0.1335
        & -- & --
        & 8.6461 & 0.1082
        & \textbf{1.1414} & \textbf{0.0442} \\

        FAST-LIO2
        & 0.8075 & \textbf{0.0213}
        & 0.3176 & 0.1161
        & \textbf{0.9861} & 0.1404
        & 0.9041 & 0.0936
        & 1.4239 & 0.1219
        & 2.8225 & 0.0595 \\

        \midrule

        \papername
        & \underline{0.7389} & \underline{0.0244}
        & \textbf{0.2245} & \textbf{0.0168}
        & 3.1403 & \underline{0.0327}
        & \underline{0.8878} & \textbf{0.0208}
        & \textbf{1.0045} & \underline{0.0235}
        & \underline{1.6376} & \underline{0.0501} \\

        \bottomrule
    \end{tabular}%
    }
\end{table*}

\begin{table*}[t]
  \centering
  \caption{Quantitative comparison on GEODE. ATE RMSE and RTE RMSE at
  $\Delta t=0.1$~s are reported in meters. All methods are evaluated on the
  same ground-truth associations within each sequence. Lower is better; best
  and second-best results are shown in bold and underlined, respectively.}
  \label{tab:geode_results}
  \setlength{\tabcolsep}{3.1pt}
  \resizebox{\textwidth}{!}{%
  \begin{tabular}{lcccccccccccccc}
    \toprule
    & \multicolumn{2}{c}{Waterways Long}
    & \multicolumn{2}{c}{Waterways Med.}
    & \multicolumn{2}{c}{Waterways Short}
    & \multicolumn{2}{c}{Offroad2}
    & \multicolumn{2}{c}{Offroad4}
    & \multicolumn{2}{c}{Offroad7}
    & \multicolumn{2}{c}{Stairs} \\
    \cmidrule(lr){2-3}\cmidrule(lr){4-5}\cmidrule(lr){6-7}
    \cmidrule(lr){8-9}\cmidrule(lr){10-11}\cmidrule(lr){12-13}
    \cmidrule(lr){14-15}
    Method
    & ATE & RTE$_{0.1}$ & ATE & RTE$_{0.1}$ & ATE & RTE$_{0.1}$
    & ATE & RTE$_{0.1}$ & ATE & RTE$_{0.1}$ & ATE & RTE$_{0.1}$
    & ATE & RTE$_{0.1}$ \\
    \midrule
    LIMOncello
    & \underline{44.8979} & \underline{0.0619}
    & \underline{14.3201} & \underline{0.0682}
    & 34.8550 & 0.0635
    & \underline{0.4449} & 0.0592
    & \underline{0.2814} & 0.0587
    & \underline{0.3636} & 0.0576
    & 3.8750 & 0.0481 \\
    RESPLE
    & 105.6537 & 0.1278
    & 137.5077 & 0.1596
    & \textbf{1.9073} & 0.0812
    & 0.4627 & 0.0643
    & 0.2968 & 0.0654
    & 0.3803 & 0.0615
    & 4.4882 & 0.0952 \\
    FAST-LIO2
    & 140.7757 & 0.1040
    & 65.9198 & 0.0901
    & \underline{30.9266} & \textbf{0.0484}
    & 0.4580 & \underline{0.0518}
    & 0.2883 & \underline{0.0539}
    & 0.3708 & \underline{0.0506}
    & \textbf{1.3639} & \textbf{0.0292} \\
    \midrule
    CT-SGAL-LIO
    & \textbf{44.8954} & \textbf{0.0420}
    & \textbf{14.3195} & \textbf{0.0387}
    & 34.8558 & \underline{0.0515}
    & \textbf{0.4398} & \textbf{0.0289}
    & \textbf{0.2798} & \textbf{0.0310}
    & \textbf{0.3608} & \textbf{0.0337}
    & \underline{3.8693} & \underline{0.0433} \\
    \bottomrule
  \end{tabular}}
\end{table*}

% The qualitative comparison on the ENWIDE
% \textit{Katzersee\_d} sequence is shown in
% Fig.~\ref{fig:enwide_katzersee_qualitative}.
% {\papername} closely follows the ground-truth trajectory over the full
% sequence, while FAST-LIO2 also maintains accurate tracking. In contrast,
% LIMOncello and RESPLE exhibit larger local trajectory deviations.
% This behavior is particularly visible in the magnified segment, where
% {\papername} remains closely aligned with the ground truth despite noticeable
% deviations in the original LIMOncello trajectory. These observations further
% support the intended role of {\papername} as a short-horizon refinement layer
% that improves local trajectory consistency without performing global
% trajectory correction.

Finally, Fig.~\ref{fig:rcampus_qualitative} examines the long-horizon
behavior of the proposed refinement. Over a 19.5-min R-Campus deployment
containing approximately 120.8k posterior states, CT-SGAL remains tightly
anchored to the underlying LIO trajectory, with only a $4.7$\, mm change in
the 3-D end-to-start gap. At the local scale, however, the reconstructed
trajectory is considerably more temporally consistent: the 95th percentile
of the consecutive-state position-rate statistic $r_k$ decreases from
$16.5$ to $2.4$\,m/s, corresponding to an $85.4\%$ reduction. This result
shows that CT-SGAL can continuously suppress posterior jumps without
introducing noticeable long-term trajectory deformation.

\subsection{Quantitative Results}
Tables~\ref{tab:grandtour_results}--\ref{tab:geode_results} report the ATE
and RTE results across GRAND-TOUR, MCD, and GEODE. Overall, {\papername}
shows its clearest advantage in short-term motion consistency. On GRAND-TOUR,
it achieves the lowest RTE on all four evaluated sequences, while also obtaining
the best ATE on Con\_3 and HAUS\_1. In particular, the RTE reductions over
LIMOncello are substantial across all four sequences, confirming that the
continuous-time refinement consistently suppresses local posterior fluctuations.

A similar trend is observed on MCD. {\papername} ranks first or second in RTE
on all six day and night sequences, while its ATE remains close to that of the
underlying LIMOncello estimator. The greatest improvements occur on
\textit{day01} and \textit{night13}, whereas on \textit{day10} and
\textit{night08} the two methods perform almost identically. This suggests that
the refinement remains stable across different operating conditions and does
not trade global accuracy for local smoothness.

On GEODE, CT-SGAL-LIO reduces RTE at $\Delta t=0.1$~s relative to
LIMOncello on all seven sequences, with reductions ranging from 9.9\% on
Stairs to 51.1\% on Offroad2. It achieves the lowest RTE on five sequences
and the second-lowest on the remaining two. ATE remains close to the input
posterior and is lowest on all three Offroad sequences and on Waterways Long
and Medium. These results match the intended behavior of CT-SGAL: the largest
change occurs in short-horizon relative motion, while the globally aligned
trajectory remains anchored to the discrete estimator.

Across the three benchmarks, the results show that {\papername} does not
systematically alter the global trajectory estimate; instead, it improves local
motion consistency while generally preserving the ATE of the discrete-time
estimator.

\subsection{Ablation Study}
\begin{table}[t]
  \centering
  \caption{GEODE ablation on nominal data. Values are geometric-mean ratios
  to the input posterior over seven sequences (input $=1$; lower is better).
  The p$_{99}$ residual is computed from a 50-ms local linear fit at 100~Hz.}
  \label{tab:geode_ablation}
  \setlength{\tabcolsep}{4.0pt}
  \begin{tabular}{lccc}
    \toprule
    Variant & ATE & RTE$_{0.1}$ & p$_{99}$ residual \\
    \midrule
    Input posterior             & 1.0000 & 1.0000 & 1.0000 \\
    No output lag ($\ell=0$)   & 0.9964 & 0.6918 & 0.0173 \\
    No curvature term           & 0.9961 & 0.6394 & 0.0081 \\
    No Huber loss               & 0.9962 & 0.6314 & 0.0035 \\
    Full CT-SGAL                & 0.9963 & 0.6370 & 0.0035 \\
    \bottomrule
  \end{tabular}
\end{table}

We ablate the output lag in Eq.~(2), the curvature term in Eq.~(7),
and the Huber loss in Eq.~(7) on the seven GEODE sequences.
The full model uses $(W,\ell)=(0.3,0.2)$~s. For the no-output-lag
variant, we set $(W,\ell)=(0.5,0)$~s such that the same $0.5$-s
history is retained. The other two variants remove the curvature
term and replace the Huber loss with ordinary least squares,
respectively.

As shown in Table~\ref{tab:geode_ablation}, the full model reduces
RTE$_{0.1}$ by 36.3\% with negligible change in ATE. The fixed output
lag yields a further 7.9\% reduction in RTE over the zero-lag variant,
while the curvature term reduces the p$_{99}$ residual by approximately
half. Removing the Huber loss changes the nominal RTE by less than
1\%, indicating that isolated posterior outliers are uncommon in these
sequences.

We further evaluate its robustness by injecting identical $0.2$-m
translation impulses into $0.5\%$ of the posterior states on three
Offroad sequences. Under these perturbations, the Huber loss reduces
the geometric-mean p$_{99}$ residual from $0.721$ to $0.482$~mm,
while ATE changes from $0.35365$ to $0.35373$~m and RTE from
$0.03048$ to $0.03110$~m. These results indicate that the Huber loss
primarily suppresses isolated posterior errors rather than affecting
nominal trajectory accuracy.
\section{Conclusions}
\label{sec:conclusions}

We presented \papername{}, a continuous-time refinement layer for the posterior pose stream of a discrete LiDAR--inertial estimator with causality. The method leaves the estimator and its map unchanged, using a history of posterior poses to fit a robust cubic trajectory in a local $SE(3)$ and evaluating it at a fixed lag so that the query lies within the observed interval.  Across 17 sequences from GRAND-TOUR, MCD, and GEODE, \papername{} mainly improves short-term motion consistency while preserving the global trajectory. It reduces RTE on all GRAND-TOUR sequences, with gains of up to 70\%, while on GEODE it achieves the best results on 10 of 14 metrics and second-best on 3, including the best ATE and RTE across all three Offroad sequences.  A 19.5-min R-Campus run also shows an 85.4\% reduction in posterior jumps with only a 4.7 mm change in the end-to-start gap.  The gains are not uniform, however, suggesting that refinement is most useful when the discrete posterior contains noticeable short-term fluctuations. Future work will test the approach with other LIO estimators and adapt the lag and fitting window to the incoming posterior's behavior.

\bibliographystyle{IEEEtran}
\bibliography{IEEEabrv, 09_references}

\end{document}